%% file: DeepQuantization_arxiv.tex
\documentclass[10pt,twocolumn,letterpaper]{article}

\usepackage{cvpr}
\usepackage{times}
\usepackage{epsfig}
\usepackage{graphicx}
\usepackage{amsmath}
\usepackage{amssymb}

\usepackage{subfigure}
\usepackage{mathrsfs}
\usepackage{amsthm}
\usepackage{amsmath}
\usepackage{amsfonts}
\usepackage{amssymb}
\usepackage{algorithm}
\usepackage{algorithmic}
\usepackage{enumerate}
\usepackage{multirow}

\input{commands.tex}

\usepackage[pagebackref=true,breaklinks=true,letterpaper=true,colorlinks,bookmarks=false]{hyperref}

\cvprfinalcopy 

\def\cvprPaperID{3107} 

\ifcvprfinal\fi
\begin{document}

\title{Deep Quantization: Encoding Convolutional Activations \\with Deep Generative Model}

\author{Zhaofan Qiu $^{\dag}$, Ting Yao $^{\ddag}$, and Tao Mei $^{\ddag}$ \\
{\small\centering$^{\dag}$University of Science and Technology of China, Hefei, China}~~~~
{\small\centering$^{\ddag}$Microsoft Research, Beijing, China}\\
{\tt\small zhaofanqiu@gmail.com, \{tiyao, tmei\}@microsoft.com}
}

\maketitle
\thispagestyle{empty}

\begin{abstract}
Deep convolutional neural networks (CNNs) have proven highly effective for visual recognition, where learning a universal representation from activations of convolutional layer plays a fundamental problem. In this paper, we present Fisher Vector encoding with Variational Auto-Encoder (FV-VAE), a novel deep architecture that quantizes the local activations of convolutional layer in a deep generative model, by training them in an end-to-end manner. To incorporate FV encoding strategy into deep generative models, we introduce Variational Auto-Encoder model, which steers a variational inference and learning in a neural network which can be straightforwardly optimized using standard stochastic gradient method. Different from the FV characterized by conventional generative models (e.g., Gaussian Mixture Model) which parsimoniously fit a discrete mixture model to data distribution, the proposed FV-VAE is more flexible to represent the natural property of data for better generalization. Extensive experiments are conducted on three public datasets, i.e., UCF101, ActivityNet, and CUB-200-2011 in the context of video action recognition and fine-grained image classification, respectively. Superior results are reported when compared to state-of-the-art representations. Most remarkably, our proposed FV-VAE achieves to-date the best published accuracy of 94.2\% on UCF101.
\end{abstract}

\section{Introduction}
The recent advances in deep convolutional neural networks (CNNs) have demonstrated high capability in visual recognition. For instance, an ensemble of residual nets~\cite{he2015deep} achieves 3.57\% in terms of top-5 error on the ImageNet dataset~\cite{russakovsky2015imagenet}. More importantly, when utilizing the activations of either a fully-connected layer or a convolutional layer in a pre-trained CNN as a universal visual representation and applying this representation to other visual recognition tasks (e.g., scene understanding, and semantic segmentation), CNNs also manifest impressive performances. The improvements are expected when CNNs are further fine-tuned with only amount of task-specific training data.

\begin{figure}[!tb]
   \centering {\includegraphics[width=0.45\textwidth]{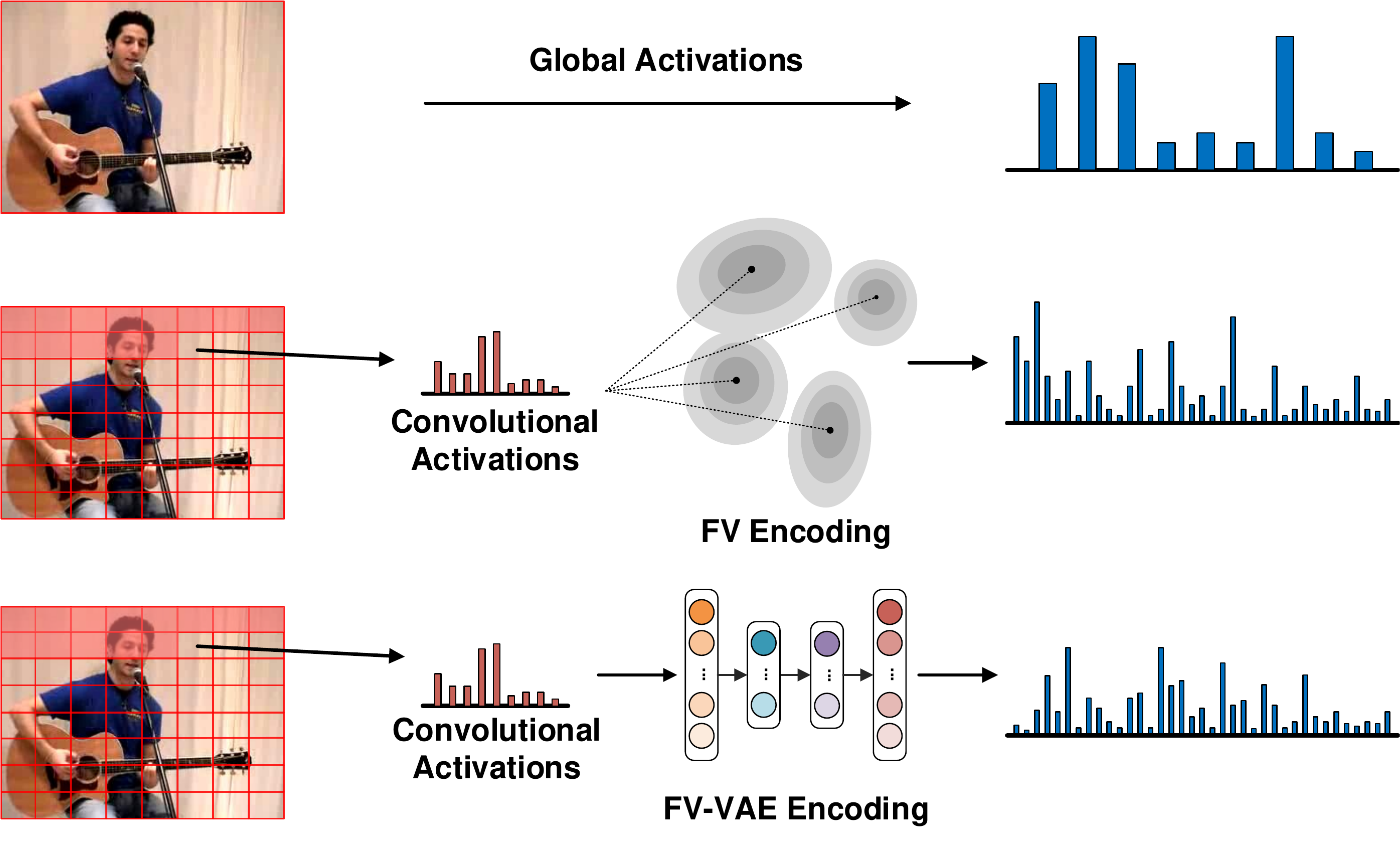}}
   \caption{\small Visual representations derived from activations of different layers in CNN (upper row: global activations of the fully-connected layer; middle row: convolutional activations with Fisher Vector encoding; bottom row: convolutional activations with our FV-VAE encoding).}
   \label{fig:fig1}
\vspace{-4mm}
\end{figure}

The activations of different layers in CNN are generally grouped into two dimensions: global activations and convolutional activations. The former directly take activations of the fully-connected layer as visual representations, which are holistic over the entire image as shown in the upper row of Figure \ref{fig:fig1}. The latter, in contrast, create visual representations by encoding a set of regional and local activations from a convolutional layer to a vectorial representation using quantization strategies, e.g., FV~\cite{perronnin2010improving} is one of the most successful quantization approaches, as illustrated in the middle row of Figure \ref{fig:fig1}. While superior results by aggregating convolutional activations are reported in most recent studies \cite{cimpoi2016deep,xu2015discriminative}, convolutional activations are first extracted as local descriptors followed by another separate quantization step. Thus such descriptors may not be optimally compatible with the encoding process, making the quantization sub-optimal. Furthermore, as discussed in \cite{johnson2016composing}, the generative model behind of FV, i.e., Gaussian Mixture Model (GMM), cannot always represent the natural clustering of the descriptors and its inflexible Gaussian observation model limits its generalization ability.

We show in this paper that these two limitations can be mitigated by designing a deep architecture for representation learning that combines convolutional activations extraction and quantization into a one-stage learning. Specifically, we present a novel Fisher Vector encoding with Variational Auto-Encoder (FV-VAE) framework to encode convolutional activations with deep generative model (i.e., Variational Auto Encoder), as shown in the bottom row of Figure \ref{fig:fig1}. The pipeline of the proposed deep architecture generally consists of two components: a sub-network with a stack of convolution layers to produce convolutional activations followed by a VAE structure aggregating the regional convolutional descriptors to a FV. VAE consists of hierarchies of conditional stochastic variables and is a highly expressive model by optimizing a variational approximation (an inference/recognition model) to the intractable posterior for the generative distribution. Compared to traditional GMM model which has the form of a mixture of fixed Gaussian components, the inference model here can be regarded as an alternative to predict specific Gaussian components to different inputs by a single neural network, making it more flexible. It is also worth noting that a classification loss is additionally considered to preserve the semantic information in the training stage. The entire architecture is trainable in an end-to-end fashion. Furthermore, in the feature extraction stage, we theoretically prove that the FV of input descriptors can be directly computed by accumulating the gradient vector of reconstruction loss in VAE through back-propagation.

The main contribution of this work is the proposal of FV-VAE architecture to encode convolutional descriptors with Variational Auto-Encoder. We theoretically formulate the computation of FV in VAE and substantiate an implementation of FV-VAE for visual representation learning.

\section{Related Work}
In the literature, visual representation generation from a pre-trained CNN model has proceeded along two dimensions: global activations and convolutional activations. The first is to extract visual representation from global activations in a CNN directly, e.g., the outputs from fully-connected layer in AlexNet \cite{krizhevsky2012imagenet} / VGG \cite{Simonyan:ICLR15} or pool5 layer in GoogleNet \cite{szegedy2015going} / ResNet \cite{he2015deep}. In practice, this scheme often starts by pre-training CNN model on a large dataset (e.g., ImageNet) and then fine-tuning the CNN architecture with a small amount of task-specific data to better characterize the intrinsic information in target scenario. The visual representation learnt in this direction has been exploited in a broad range of computer vision tasks including fine-grained image classification~\cite{branson2014bird,krause2015fine}, video action recognition~\cite{karpathy2014large,simonyan2014two} and visual captioning~\cite{venugopalan2015sequence,vinyals2015show}.

Another alternative scheme is to utilize the activations from convolutional layers in CNN as regional and local descriptors. Compared to global activations, convolutional activations from CNN are embedded with rich spatial information, making them more transferable to different domains and more robust to translation and rotation, which have shown the effectiveness in several technological advances, e.g., Spatial Pyramid Pooling~(SPP)~\cite{he2014spatial}, Fast R-CNN~\cite{girshick2015fast} and Fully Convolutional Networks~(FCNs)~\cite{long2015fully}. Recently, many works attempt to produce visual representation by encoding convolutional activations with different quantization strategies. For example, Fisher Vector \cite{perronnin2010improving} is computed on the output of the last convolutional layer of VGG networks for describing texture in \cite{cimpoi2016deep}. Similar in spirit, Xu \emph{et al.} utilize VLAD \cite{jegou2010aggregating} to encode convolutional descriptors of video frame for multimedia event detection \cite{xu2015discriminative}. In \cite{sharma2015action} and \cite{xu2015show}, Sharma \emph{et al.} and Xu \emph{et al.} dynamically pool convolutional descriptors with attention models for action recognition and image captioning, respectively. Furthermore, convolutional descriptors of one convolutional layer are pooled with the guidance of the activations of the successive convolutional layer in \cite{liu2015treasure}. In \cite{gao2015compact} and \cite{lin2015bilinear}, convolutional descriptors from two CNNs are multiplied using the outer product and pooled to obtain the bilinear vector.

In summary, our work belongs to the second dimension and aims to compute FV on convolutional activations with deep generative models. We exploit Variational Auto-Encoder for this purpose, which optimizes an inference model to the intractable posterior. The high flexibility of the inference model and efficiency of the structure optimization makes VAE more advanced than traditional GMM. Our work in this paper contributes by studying not only encoding convolutional activations in a deep architecture, but also theoretically figuring out the computation of FV based on VAE architecture.

\section{Fisher Vector Meets VAE}
In this section, we first recall the Fisher Vector theory, followed by presenting how to estimate the probability density function in FV through VAE. The optimization of VAE is then elaborated and how to compute the FV of the input descriptors will be introduced finally.

\subsection{Fisher Vector Theory} \label{sec:FVT}
Suppose we have two sets of local descriptors $X=\{\bxt\}^{T_x}_{t=1}$ and $Y=\{\byt\}^{T_y}_{t=1}$ with ${T_x}$ and $T_y$ descriptors, respectively. Let $\bxt,\byt\in\mathbb{R}^d$ denote the $d$-dimensional features of each descriptor. In order to measure the similarity between the two sets, kernel method is employed by mapping them into a hyperspace. Specifically, assuming that the generation process of descriptors in $\mathbb{R}^d$ can be modeled by a probability density function $\uTheta$ with $M$ parameters $\bTheta=[{\theta}_{1},...,{\theta}_{M}]'$, Fisher Kernel (FK)~\cite{jaakkola1999exploiting} between the two sets $X$ and $Y$ is given by
\begin{equation}\small
\begin{aligned}
K(X, Y)={G_{\bTheta}^{X}}'F_{\bTheta}^{-1}G_{\bTheta}^{Y}
\end{aligned}~~,
\label{equ:FK}
\end{equation}
where $G_{\bTheta}^{X}={\nabla}_{\bTheta}\log{\uTheta}(X)$ is defined as fisher score function by computing the gradient of the log-likelihood of the set based on the generative model, and $F_{\bTheta}=\mathbb{E}_{X\sim \uTheta}[G_{\bTheta}^{X}{G_{\bTheta}^{X}}']$ is the Fisher Information Matrix (FIM) of $\uTheta$ which is regarded as statistical feature normalization.
Since $F_{\bTheta}$ is positive semi-definite, the FK in Eq.(\ref{equ:FK}) can be re-written explicitly as inner product in hyperspace:
\begin{equation}\small
\begin{aligned}
K(X, Y)={\mathscr{G}_{\bTheta}^{X}}'\mathscr{G}_{\bTheta}^{Y}
\end{aligned}~~,
\end{equation}
where
\begin{equation}\small
\begin{aligned}
\mathscr{G}_{\bTheta}^{X}=F_{\bTheta}^{-\half}G_{\bTheta}^{X}=F_{\bTheta}^{-\half}{\nabla}_{\bTheta}\log{\uTheta}(X)
\end{aligned}~~.
\label{equ:FV}
\end{equation}
Formally, $\mathscr{G}_{\bTheta}^{X}$ is well-known as Fisher Vector (FV). The dimension of FV is equal to the number of generative parameters $\bTheta$, which is often much higher than that of the descriptor, making FV of higher descriptive capability.

\begin{figure}[!tb]
   \centering
   \subfigure[VAE Training]{
     \label{fig:fig2:a}
     \includegraphics[width=0.49\textwidth]{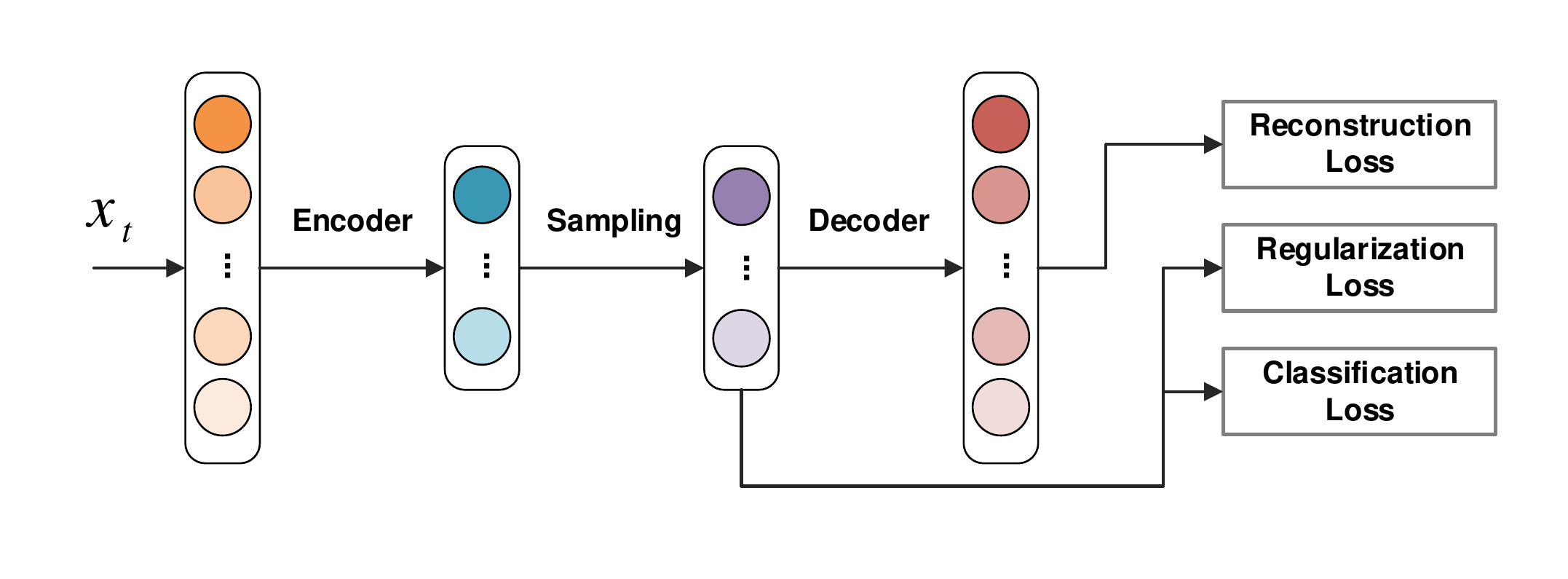}}
   \subfigure[FV Extraction]{
     \label{fig:fig2:b}
     \includegraphics[width=0.49\textwidth]{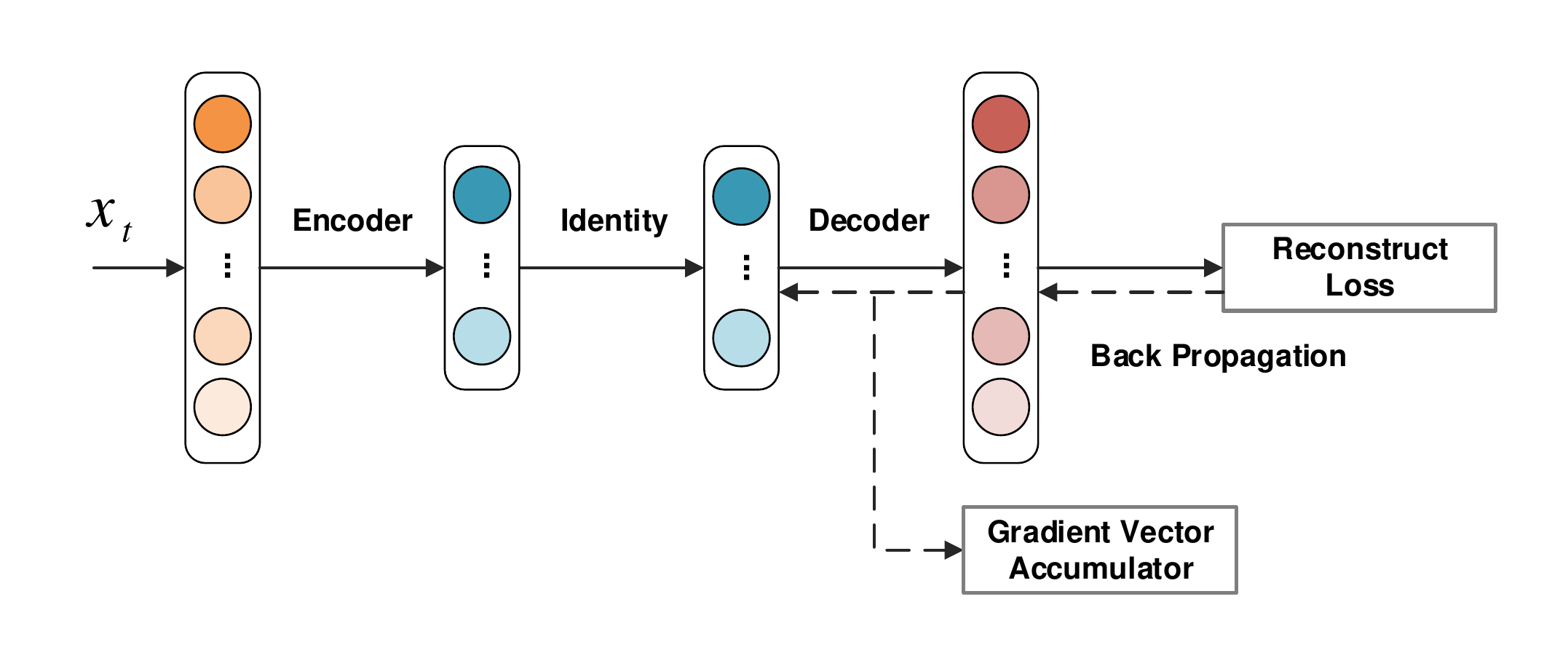}}
   \caption{\small The overview of FV learning with VAE: (a) The training process of VAE, (b) FV extraction based on VAE.}
   \label{fig:fig2}
   \vspace{-0.1in}
\end{figure}

\subsection{Probability Estimation through VAE}
Next, we will discuss how to estimate the probability density function $\uTheta$ in FV. In general, $\uTheta$ is chosen to be Gaussian Mixture Model (GMM)~\cite{sanchez2013image,wang2013action} as one can approximate any distribution with arbitrary precision by GMM, in which $\bTheta$ is composed of mixture weight, mean and covariance of Gaussian components. The need of a large number of mixture components and inefficient optimization of Expectation Maximization algorithm, however, makes the parameter learning computationally expensive and difficult to be applied to large-scale complex data. Inspired by the idea of deep generative models \cite{kingma2013auto,rezende2014stochastic} which enable the flexible and efficient inference learning in a neural network, we develop a Variational Auto-Encoder (VAE) to generate the probability function $\uTheta$.

Following the notations in Section~\ref{sec:FVT} and assuming that all the descriptors in the set are independent, the log-likelihood of the set can be calculated by the sum over log-likelihoods of individual descriptor and written as
\begin{equation}\small
\begin{aligned}
\log{\uTheta}(X)={\sum_{t=1}^{T_x}}{\log{\pTheta(\bxt)}}
\end{aligned}~~.
\end{equation}

To model the probability of $\bxt$ generated from parameters $\bTheta$, an unobserved continuous random variable $\bz_t$ is involved with prior distribution $\pTheta(\bz)$ and each $\bxt$ is generated from the conditional distribution $\pTheta(\mathbf{x}|\bz)$. As such, each log-likelihood $\log{\pTheta(\bxt)}$ can be measured as
\begin{equation}\small
\begin{aligned}
\log{\pTheta(\bxt)}&=D_{KL}(\qPhi(\bz|\bxt)||\pTheta(\bz|\bxt))+\mathcal{LB}(\bTheta,\bPhi;\bxt) \\
&\geqslant \mathcal{LB}(\bTheta,\bPhi;\bxt)
\end{aligned}~,
\end{equation}
where $\mathcal{LB}(\bTheta,\bPhi;\bxt)$ is the variational lower bound on the likelihood of descriptor $\bxt$ and can be written as
\begin{equation}\small
\mathcal{LB}(\bTheta,\bPhi;\bxt) = -D_{KL}(\qPhi(\bz|\bxt)||\pTheta(\bz))+\mathbb{E}_{\qPhi(\bz|\bxt)}[\log{\pTheta(\bxt|\bz)}],
\label{equ:LB}
\end{equation}
where $\qPhi(\bz|\mathbf{x})$ is a recognition model which is an approximation to the intractable posterior $\pTheta(\bz|\bx)$. In our proposed FV-VAE method, we use this lower bound $\mathcal{LB}(\bTheta,\bPhi;\bxt)$ as an approximation of the log-likelihood. Through this approximation, the generative model can be divided into two parts: encoder $\qPhi(\bz|\mathbf{x})$ and decoder $\pTheta(\bx|\bz)$, predicting hidden and visible probability, respectively.

\begin{algorithm}[!tb]\small
\caption{\small Variational Auto-Encoding (VAE) Optimization}\label{alg:vae}
\begin{algorithmic}[1]
    \STATE \textbf{Input:}
        training set $X=\{\bxt\}^{T_x}_{t=1}$, corresponding labels $L=\{l_t\}^{T_x}_{t=1}$, loss weights ${\lambda}_{1}, {\lambda}_{2}, {\lambda}_{3}$.
    \STATE \textbf{Initialization:}
        random initialized $\bTheta_0,\bPhi_0$.
    \STATE \textbf{Output:}
        VAE parameters $\bTheta^{\ast},\bPhi^{\ast}$.
    \REPEAT
    \STATE
        Sample $\bxt$ in the minibatch.
    \STATE
        Encoder: $\bMu_{\bz_t} \leftarrow f_{\bPhi}(\bxt)$.
    \STATE
        Sampling: ${\bz_t} \leftarrow \bMu_{\bz_t}+\bEpsilon\odot \bSigma_{\bz}, \bEpsilon\sim \mathcal{N}(0,\mathbf{I})$.
    \STATE
        Decoder: $\bMu_{\bxt} \leftarrow f_{\bTheta}(\bz_t)$.
    \STATE
        Compute reconstruction loss: \\
        ~~~~$\mathcal{L}_{rec}=-\log{\pTheta(\bxt|\bz_t)}=-\log \mathcal{N}(\bxt;\bMu_{\bxt},\bSigma_{\mathbf{x}}^2\mathbf{I})$. \\
    \STATE
        Compute regularization loss: \\
        ~~~~$\mathcal{L}_{reg}=\half \left \| {\bMu_{\bz_t} } \right \|+\half \left \| {\bSigma_{\bz} } \right \|-\half\sum_{k=1}^{d}(1+\log {\sigma}_{\bz,d}^{2})$. \\
    \STATE
        Compute classification loss: \\
        ~~~~$\mathcal{L}_{cls}=softmax\_loss(\bz_t, l_t)$.
    \STATE
        Fuse the three loss: \\
        ~~~~$\mathcal{L}(\bTheta,\bPhi)={\lambda}_{1}\mathcal{L}_{rec}(\bTheta,\bPhi)+{\lambda}_{2}\mathcal{L}_{reg}(\bPhi)+{\lambda}_{3}\mathcal{L}_{cls}(\bPhi)$.
    \STATE
        Back-propagate the gradients.
    \UNTIL{maximum iteration reached.}
\end{algorithmic}
\end{algorithm}

\begin{figure*}[!tb]
   \centering {\includegraphics[width=0.96\textwidth]{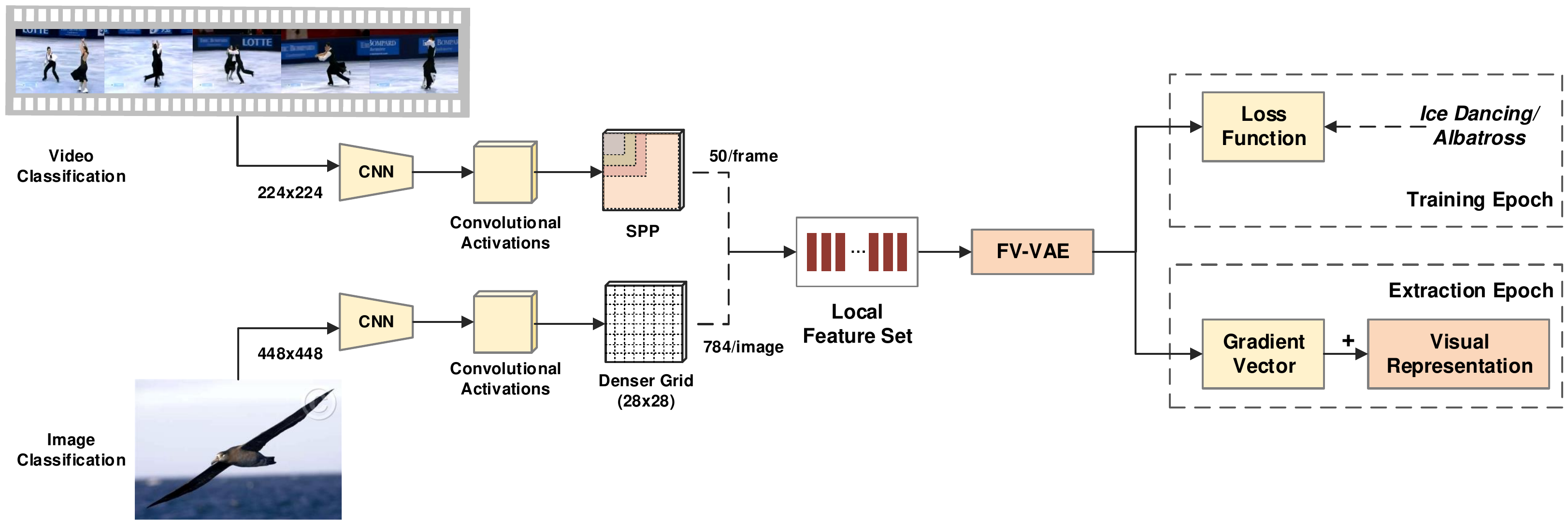}}
   \caption{\small Visual representation learning framework for image and video recognition. Spatial Pyramid Pooling (SPP) is performed on the last pooling layer of CNN to aggregate the local descriptors of video frame, which applies four different max pooling operations and obtain $(6\times6)$, $(3\times3)$, $(2\times2)$ and $(1\times1)$ outputs for each convolutional filter, resulting a total of 50 descriptors. For image, an input with a higher resolution of ($448\times 448$) is fed into the CNN and the activations of the last convolutional layer conv$_{5\_4}$+relu in VGG\_19 are extracted, leading to dense local descriptors of $28\times 28$ for VGG\_19. In training stage, FV-VAE architecture is learnt by minimizing the overall loss. In extraction epoch, the learnt FV-VAE is to encode the set of local descriptors into a vectorial FV representation.}
   \label{fig:fig3}
   \vspace{-0.15in}
\end{figure*}

\subsection{VAE Optimization}
The inference model parameter $\bPhi$ and generative model parameter $\bTheta$ are straightforward to be optimized using stochastic gradient descend method. More specifically, let the prior distribution be the standard normal distribution $\pTheta(\bz)=\mathcal{N}(\bz;0,\mathbf{I})$, and both the conditional distribution $\pTheta(\bx|\bz)$ and posterior approximation $\qPhi(\bz|\mathbf{x})$ be multivariate Gaussian distribution $\mathcal{N}(\bxt;{\bMu}_{\bxt},{\bSigma}_{\bxt}^2\mathbf{I})$ and $\mathcal{N}(\bzt;{\bMu}_{\bz_t},{\bSigma}_{\bz_t}^2\mathbf{I})$, respectively. The one-step Monte Carlo is exploited to estimate the latent variable $\bz_t$. Hence, the lower bound in Eq.~(\ref{equ:LB}) can be rewritten as
\begin{equation}\small
\begin{aligned}
\mathcal{LB}(\bTheta,\bPhi;\bxt)\simeq & \log{\pTheta(\bxt|\bz_t)}+\half\sum_{k=1}^{d}(1+\log {\sigma}_{\bz_t,d}^{2}) \\
& -\half \left \| {\bMu_{\bz_t} } \right \|-\half \left \| {\bSigma_{\bz_t} } \right \|
\end{aligned}~~,
\label{equ:LB2}
\end{equation}
where $\bz_t$ is generated from $\mathcal{N}(\bMu_{\bz_t}, \bSigma_{\bz_t}^2\mathbf{I})$ and it is equivalent to $\bz_t=\bMu_{\bz_t}+ \bEpsilon\odot \bSigma_{\bz_t}$,$\bEpsilon\sim \mathcal{N}(0,\mathbf{I})$.

Figure \ref{fig:fig2:a} illustrates an overview of our VAE training process and Algorithm \ref{alg:vae} further details the optimization steps. It is also worth noting that different from the training of standard VAE method which estimates $\bSigma_{\mathbf{x}}$ and $\bSigma_{\bz}$ in another parallel encoder-decoder structure, we simply learn the two covariance by gradient descent technique and share them across all the descriptors, making the number of parameters learnt in VAE significantly reduced in our case. In addition to the basic reconstruction loss and regularization loss, we further take classification loss into account in our VAE training to incorporate semantic information, which has been shown effective in semi-supervised generative model learning \cite{kingma2014semi}. The overall loss function is then given by
\begin{equation}\small
\begin{aligned}
\mathcal{L}(\bTheta,\bPhi)={\lambda}_{1}\mathcal{L}_{rec}(\bTheta,\bPhi)+{\lambda}_{2}\mathcal{L}_{reg}(\bPhi)+{\lambda}_{3}\mathcal{L}_{cls}(\bPhi)
\end{aligned}~~.
\label{equ:LF}
\end{equation}
We fix ${\lambda}_{1}={\lambda}_{2}=1$ in Eq.~\eqref{equ:LF} and will investigate the effect of tradeoff parameter ${\lambda}_{3}$ in our experiments. During the training, the gradients are calculated and back-propagate to the lower layers so that lower layers can adjust their parameters to minimize the loss.

\subsection{FV Extraction} \label{sec:FVEX}
After the optimization of model parameters $[\bTheta^{\ast},\bPhi^{\ast}]$, Figure~\ref{fig:fig2:b} demonstrates how to extract Fisher Vector based on the learnt VAE architecture.

By replacing the log-likelihood with its approximation, i.e., lower bound $\mathcal{LB}(\bTheta,\bPhi;\bxt)$, we can obtain FV in Eq.~\eqref{equ:FV}:
\begin{equation}
\small
\begin{aligned}
\mathscr{G}_{\bTheta^{\ast}}^{X}&=F_{\bTheta^{\ast}}^{-\half} {\nabla}_{\bTheta}\log{u_{\bTheta^{\ast}}}(X)\\
&=-F_{\bTheta^{\ast}}^{-\half} \sum_{t=1}^{T_x}[{\nabla}_{\bTheta}\mathcal{L}_{rec}(\bxt; \bTheta^{\ast},\bPhi^{\ast})]
\end{aligned}~~,
\label{equ:FV-VAE}
\end{equation}
which is the normalized gradient vector of reconstruction loss, and can be computed directly though the back propagation operation. It is worth noticing that when extracting FV representation, we withdraw the sampling operation and use $\bMu_{\bz_t}$ as $\bz_t$ directly to avoid stochastic factors.

\section{Visual Representation Learning}
By utilizing FV-VAE as a deep architecture for quantization, a general visual representation learning framework is devised for image and video recognition, respectively, as illustrated in Figure~\ref{fig:fig3}. The basic idea is to construct a set of convolutional descriptors for image or video frames, followed by encoding them into a vectorial FV representation using FV-VAE architecture. Both the training epoch and FV extraction epoch are shown in Figure~\ref{fig:fig3} and the entire framework is trainable in an end-to-end manner.

We exploit different strategies of aggregation to construct the set of convolutional descriptors for video frames and image, respectively, due to the different property in between. A video consists of a sequence of frames with large intra-class variations caused by, e.g., camera motion, illumination conditions and so on, making the scale of an identical object varying in different frames. Following \cite{xu2015discriminative}, we employ Spatial Pyramid Pooling (SPP)~\cite{he2014spatial} on the last pooling layer to extract scale-invariant local descriptors for video frames. Instead, we feed a higher resolution (e.g., $448\times 448$) input into the CNN to fully utilize image information and extract the activations of the last convolutional layer (e.g., conv$_{5\_4}$+relu in VGG\_19), resulting in dense local descriptors (e.g., $28\times 28$) for image as in~\cite{lin2015bilinear}.

In our implementation, Multi-Layer Perceptron (MLP) is employed as encoder and decoder in FV-VAE and one layer decoder is developed to reduce the dimension of FV representation. As such, the functions in Algorithm~\ref{alg:vae} can be specified as
\begin{equation}
\small
\begin{aligned}
~~~~~&\textbf{Encoder}: \bMu_{\bz_t} \leftarrow {MLP}_{\bPhi}(\bxt)\\
~~~~~&\textbf{Decoder}: \bMu_{\bxt} \leftarrow ReLU({W_{\bTheta}'}\bzt+\bb{b}_{\bTheta})\\
\end{aligned}~~,
\end{equation}
where \{$W_{\bTheta}$, $\bb{b}_{\bTheta}$\} are the encoder parameters $\bTheta$. The gradient vector of $\mathcal{L}_{rec}$ is calculated as
\begin{equation}
\small
\begin{aligned}
{\nabla}_{\bTheta}\mathcal{L}_{rec}(\bxt; \bTheta^{\ast},\bPhi^{\ast}) &= flatten\left\{[\frac{\partial \mathcal{L}_{rec}}{\partial W_{\bPhi}},\frac{\partial \mathcal{L}_{rec}}{\partial \bb{b}_{\bTheta}}]\right\} \\
&=flatten\left\{[\frac{\partial \mathcal{L}_{rec}}{\partial \bMu_{\bxt}}\cdot \bzt',\frac{\partial \mathcal{L}_{rec}}{\partial \bMu_{\bxt}}] \right\} \\
&=flatten\left\{\frac{\partial \mathcal{L}_{rec}}{\partial \bMu_{\bxt}}\cdot [\bzt',1]\right\} \\
&=flatten\left\{\frac{\bMu_{\bxt}-\bxt}{\bSigma_{\mathbf{x}}^2}\odot{(\bMu_{\bxt}>0)}\cdot [\bzt',1]\right\}
\end{aligned},
\end{equation}
where ``$flatten$'' represents to flatten a matrix to a vector, and $\odot$ denotes element-wise multiplication to filter the activated elements. Considering it is difficult to obtain an analytic solution of FIM in this case, we make an approximation by replacing the expectation with the average on the whole training set:
\begin{equation}
\small
\begin{aligned}
F_{\bTheta^{\ast}}=\mathbb{E}_{X\sim \uTheta}[G_{\bTheta}^{X}{G_{\bTheta}^{X}}']\approx\mathop{mean}\limits_{X}[G_{\bTheta}^{X}{G_{\bTheta}^{X}}']
\end{aligned}~~,
\end{equation}
and
\begin{equation}
\small
\begin{aligned}
\mathscr{G}_{\bTheta^{\ast}}^{X}=flatten\left\{-F_{\bTheta^{\ast}}^{-\half} \cdot \sum_{t=1}^{T_x}(\frac{\bMu_{\bxt}-\bxt}{\bSigma_{\mathbf{x}}^2}\odot{(\bMu_{\bxt}>0)}\cdot [\bzt',1])\right\}
\label{equ:FV-VAE-FINAL}
\end{aligned}~~,
\end{equation}
which is the output FV representation in our framework.

To improve the convergence speed and better regularize the visual representation learning for video, we train this framework by inputting one single video frame rather than multiple ones, which is randomly sampled from videos. In the FV extraction stage, the video-level representation can be easily obtained by averaging FVs of all the frames sampled from the video since FV in Eq.~(\ref{equ:FV-VAE-FINAL}) is linear additive.

\section{Experiments}

We evaluate the learnt visual representation by FV-VAE architecture on three popular datasets, i.e., UCF101~\cite{UCF101}, ActivityNet~\cite{caba2015activitynet} and CUB-200-2011~\cite{WahCUB_200_2011}. The UCF101 dataset is one of the most popular video action recognition benchmarks. It consists of 13,320 videos from 101 action categories. The action categories are divided into five groups: human-object interaction, body-motion only, human-human interaction, playing musical instruments and sports. Three training/test splits are provided by the dataset organisers and each split in UCF101 includes about 9.5K training and 3.7K test videos. The ActivityNet dataset is a large-scale video benchmark for human activity understanding. The latest released version of the dataset (v1.3) is exploited, which contains 19,994 videos from 200 activity categories. The 19,994 videos are divided into 10,024, 4,926, 5,044 videos for training, validation and test set, respectively. Note that the labels of test set are not publicly available and the performances on ActivityNet dataset are all reported on validation set. Furthermore, we also validate the representation on CUB-200-2011 dataset, which is widely adopted for fine-grained image classification and consists of 11,788 images from 200 bird species. We follow the official split on this dataset with 5,994 training and 5,794 test images.

\begin{table}
\centering
\small
\caption{\small Methodology comparison of different quantization.}
\begin{tabular}{|l|p{0.33\columnwidth}|p{0.36\columnwidth}|}   \hline
Quantization & indicator & descriptor\\ \hline\hline
FV~\cite{perronnin2010improving} & Gaussian observation model & gradient with respect to GMM parameters\\ \hline
VLAD~\cite{jegou2010aggregating} & clustering center & difference to the assigned center\\ \hline
BP~\cite{lin2015bilinear} & local feature & coordinate representation\\ \hline
FV-VAE & VAE hidden variable & gradient of reconstruction loss\\ \hline
\end{tabular}
\label{table:theoretical}
\vspace{-2mm}
\end{table}

\subsection{Compared Approaches}
To empirically verify the merit of visual representation learnt by FV-VAE, we compare the following quantization methods: \textbf{Global Activations (GA)} directly utilizes the outputs of fully-connected/pooling layer as visual representation. \textbf{Fisher Vector (FV)}~\cite{perronnin2010improving} produces the visual representation by concatenating the gradients with respect to the parameters of GMM, which is trained on local descriptors. \textbf{Vector of Locally Aggregated Descriptors (VLAD)}~\cite{jegou2010aggregating} is to accumulate, for each clustering center learnt with K-means, the differences between the clustering center and the descriptors assigned to it, and then concatenates the accumulated vector of each center as quantized representation. \textbf{Bilinear Pooling (BP)}~\cite{lin2015bilinear} pools local descriptors in a pairwise manner by outer product. In our case, one local descriptor pairs with itself. To better illustrate the difference between the compared approaches, we details the methodology in Table~\ref{table:theoretical}. In particular, we decouple the quantization process into two parts: indicator and descriptor. Indicator refers to observations/distributions estimated on the whole set of local descriptors and descriptor is to represent the set with respect to the indicator.

\subsection{Experimental Settings}
\textbf{Convolutional activations.} On video action recognition task, we extract two widely adopted convolutional activations, i.e., activations of pool5 layer in VGG\_19~\cite{Simonyan:ICLR15} and res5c layer in ResNet\_152~\cite{he2015deep}. Given a $224\times 224$ video frame as input, the outputs of the two layers are both $7\times7$ and the dimension of each activation is 512 and 2,048, respectively. For each video, 25 frames are uniformly sampled for representation extraction. On image classification problem, we feed $448\times 448$ image into VGG\_19 and the activations of conv$_{5\_4}$+relu layer are exploited, which produce $28\times 28$ convolutional descriptors.

\textbf{VAE optimization.} To make the training process of VAE stable, we first exploit L2 normalization on each convolutional activation to make the input to VAE in a common scale. Following~\cite{Im2016dvae,vincent2008extracting}, dropout is then employed to randomly drop out units input to the encoder but the auto-encoder is optimized to reconstruct a complete ``repaired" input. The dropout rate is fixed to 0.5. Furthermore, we utilize AdaDelta~\cite{zeiler2012adadelta} optimization method implemented in Caffe~\cite{jia2014caffe} to normalize the gradient of each parameters for balancing their converge speed. The base learning rate is set to 1 and the size of mini-batch is 128 images/frames. The optimization will be complete after 5,000 batches.

\textbf{Quantization settings.} For our FV-VAE, given the local descriptor with dimension $C$ ($C\in \{512,2048\}$), we design a two-layer encoder ($C\rightarrow C\rightarrow 255$) to reduce the dimension to 255, coupled with a single layer decoder ($255\rightarrow C$). The dimension of the final quantized representation is $256\times C$. For FV and VLAD, we follow the settings in~\cite{cimpoi2016deep} and~\cite{xu2015discriminative}. Specifically, 128 Gaussian components for FV and 256 clustering centers for VLAD are exploited. As such, the dimension of representations encoded by FV and VLAD will also be $256\times C$. The two quantization approaches are implemented by VLFeat~\cite{vedaldi08vlfeat}.

\textbf{Classifier training.} After representation learning by all the methods in our experiments, we apply signed square-root step ($sign(x)\sqrt{\left | x \right |}$) and L2 normalization ($x/{\left \| x \right \|}_2$) as in \cite{cimpoi2016deep,lin2015bilinear,perronnin2010improving,xu2015discriminative}, and then train a one-vs-all linear SVM with a fixed hyperparameter $C_{svm}=100$.

\begin{table}
\centering
\small
\caption{\small Performance comparisons of different quantization methods on UCF101 split1 with default VGG\_19 network.}
\begin{tabular}{|l|l|*{2}{c|}} \hline
Feature                   & Dimension & Accuracy \\ \hline \hline
GA               & 4096      & 74.91\% \\ \hline
Concatenation    & 25088     & 75.89\% \\
AVE                 & 512       & 73.25\%\\ \hline
FV                  & 131072     & 78.85\% \\
VLAD                & 131072    & 80.67\% \\
BP                 & 262144     & 81.39\% \\ \hline
FV-VAE$^-$                & 131072    & 81.91\% \\
FV-VAE      & 131072    & 83.45\% \\ \hline
\end{tabular}
\label{tab:quantizations}
\vspace{-4mm}
\end{table}

\subsection{Performance Comparison}
\textbf{Comparison with different quantization methods.} We first examine our FV-VAE and compare with other quantization methods. In addition to the four mentioned quantization methods, we also include three runs: Concatenation, AVE and FV-VAE$^-$. Concatenation is to flatten the activations of pool5 layer and concatenate into a super vector, whose dimension is 25088 (7 $\times$ 7 $\times$ 512). The representation in AVE is produced by average fusing the 49 512-dimensional convolutional activations in pool5 layer. A slightly different setting of our FV-VAE is named as FV-VAE$^-$, in which the classification loss in Eq.(\ref{equ:LF}) is excluded or ${\lambda}_{3}$ is set to 0.

The performances and comparisons with default VGG\_19 network on UCF101 (split 1) are summarized in Table \ref{tab:quantizations}. Overall, the results indicate that our FV-VAE leads to a performance boost against others. In particular, the accuracy of FV-VAE can achieve 83.45\%, which makes the relative improvement over the best competitor BP by 2.5\%. Meanwhile, the dimension of representation learnt by FV-VAE is only half of that of BP. There is a performance gap among three runs GA, Concatenation and AVE. Though three runs all directly originate from pool5 layer, they are fundamentally different in the way of generating frame representation. The representation of GA is as a result of flatting all kernel maps in pool5 to the neurons in a fully-connected layer, while Concatenation and AVE is by directly concatenating convolutional descriptors or average fusing them in pool5 layer. As indicated by our results, Concatenation can lead to better performance than GA and AVE. VLAD outperforms FV on UCF101, but the performance is still lower than BP. Compared to FV which produces representation with respect to a number of Gaussian mixture components, FV-VAE will learn which Gaussian distribution is needed for the input specific descriptor by an inference neural network, making FV-VAE more flexible. Therefore, FV-VAE performs significantly better than FV. More importantly, FV-VAE is trainable in an end-to-end fashion. By additionally incorporating semantic information, FV-VAE leads to apparent improvement against FV-VAE$^-$. Furthermore, by reducing the dimension of latent variable to 7, the visual representations produced by FV-VAE and GA are then with the same dimension of 4,096. In this case, the accuracy of FV-VAE can still achieve 78.37\% which is higher than 74.91\% by GA, again demonstrating the effectiveness of our FV-VAE. In addition, similar performance trends are observed at CUB-200-2011 dataset, as shown in the upper rows of Table~\ref{tab:cub}, in two protocols of where the object bounding boxes are provided or not.

\begin{table}
\centering
\small
\caption{\small Performance comparisons of FV-VAE with local activations from different networks on UCF101 split1.}
\begin{tabular}{|l|c|c|c|} \hline
Network                     & \multicolumn{1}{c|}{~~~~~~GA~~~~~~}           & \multicolumn{1}{c|}{FV-VAE$^-$}  & \multicolumn{1}{c|}{~~FV-VAE~~} \\ \hline \hline
pool5                     & 74.91\%       & 81.91\%                   & 83.45\% \\ \hline
pool5 fine-tuned           & 79.06\%       & 82.05\%                   & 82.13\% \\ \hline
res5c                 & 81.57\%       & 85.05\%                   & 86.33\% \\ \hline
\end{tabular}
\label{tab:networks}
\vspace{-1mm}
\end{table}

\begin{table}
\centering
\small
\caption{\small Performance comparisons with the state-of-the-art methods on UCF101 (3 splits, $\times$10 augmentation). C3D: Convolutional 3D \cite{tran2015learning}; TSN: Temporal Segment Networks; TDD: Trajectory-pooled Deep-convolutional Descriptor \cite{wang2015action}; IDT: Improved Dense Trajectory \cite{wang2013action}.}
\begin{tabular}{|l|*{1}{c|}} \hline
Method                                                      & Accuracy \\ \hline \hline
Two-stream ConvNet \cite{simonyan2014two}                   & 88.1\% \\
C3D (3 nets) \cite{tran2015learning}                        & 85.2\% \\
Factorized ST-ConvNet \cite{sun2015human}                   & 88.1\% \\
Two-stream + LSTM \cite{yue2015beyond}                      & 88.6\% \\
Two-stream fusion \cite{feichtenhofer2016convolutional}     & 92.5\% \\
Long-term temporal ConvNet \cite{varol2016long}             & 91.7\% \\
Key-volume mining CNN \cite{zhu2016key}                     & 93.1\% \\
TSN (3 modalities) \cite{wang2016temporal}                  & 94.2\% \\ \hline
IDT \cite{wang2013action}                                   & 85.9\% \\
C3D + IDT \cite{tran2015learning}                           & 90.4\% \\
TDD + IDT \cite{wang2015action}                             & 91.5\% \\
Long-term temporal ConvNet + IDT \cite{varol2016long}       & 92.7\% \\ \hline
FV-VAE-pool5                                                & 83.9\% \\
FV-VAE-pool5 optical flow                                   & 89.5\% \\
FV-VAE-res5c                                            & 86.6\% \\ \hline
FV-VAE-(pool5 + pool5 optical flow)                       & 93.7\% \\
FV-VAE-(res5c + pool5 optical flow)                   & 94.2\% \\
FV-VAE-(res5c + pool5 optical flow) + IDT             & 95.2\% \\ \hline
\end{tabular}
\label{tab:ucf101}
\vspace{-4mm}
\end{table}

\textbf{Comparison with different networks.} Next, we turn to measure the performance comparison on UCF101 split1 of our FV-VAE with local activations from different networks, including pool5 layer in VGG\_19 and fine-tuned VGG\_19 using video frames respectively, and res5c layer in ResNet\_152. Compared to pool5 in VGG\_19, FV-VAE on the outputs of res5c layer in ResNet\_152 with a deeper CNN exhibits better performance. An interesting observation is that GA and FV-VAE$^-$ performs better on the outputs of pool5 layer in fine-tuned VGG\_19 than that in VGG\_19, while reverse trend is indicated by using FV-VAE. We speculate that this may be the result of overfitting in fine-tuning with UCF101, which in particular affects the descriptive ability of convolutional layers. This result also indicates the advantage of exploring semantic information in FV-VAE training based on the outputs of a general network than a fine-tuned one.

\textbf{Comparison with the state-of-the-art.} We compare with several state-of-the-art techniques on three splits of UCF101, ActivityNet validation set and CUB-200-2011. The performance comparisons are summarized in Table \ref{tab:ucf101}, \ref{tab:activitynet} and \ref{tab:cub}, respectively. It is worth noting that most recent works on UCF101 employ and fuse two or multiple modalities. For fair comparison, two basic and widely adopted modalities, i.e., video frame and optical flow ``image," are considered as inputs to our visual representation framework and late fusion is used to combine classifier scores on the two modalities. As shown in Table \ref{tab:ucf101}, FV-VAE on activations from pool5 layer in VGG\_19 with image and optical flow inputs can achieve 93.7\%, which makes the relative improvement over two-stream networks \cite{simonyan2014two}, \cite{yue2015beyond} and \cite{feichtenhofer2016convolutional} by 6.3\%, 5.7\% and 1.3\%, respectively. When exploiting the outputs of res5c layer in ResNet\_152 on image inputs as instead, the accuracy will be further improved to 94.2\%. By combining with IDT which are hand-crafted features, our final performance will boost up to 95.2\%, which is to-date the best published performance on UCF101.

\begin{table}
\centering
\small
\caption{\small Performance comparisons in terms of Top-1\&Top-3 classification accuracy, and mean AP on ActivityNet validation set.}
\begin{tabular}{|l|*{4}{c|}} \hline
Methods                             & Top-1       & Top-3      & MAP \\ \hline \hline
VGG\_19-GA \cite{Simonyan:ICLR15}      & 66.59\%    & 82.70\%   & 70.22\% \\
ResNet\_152-GA \cite{he2015deep}       & 71.43\%    & 86.45\%   & 76.56\% \\
C3D-GA \cite{tran2015learning}         & 65.80\%    & 81.16\%   & 67.68\% \\
IDT \cite{wang2013action}           & 64.70\%    & 77.98\%   & 68.69\% \\ \hline
FV-VAE-pool5                        & 72.51\%    & 85.68\%   & 77.25\% \\
FV-VAE-res5c                        & 78.55\%    & 91.16\%   & 84.09\% \\ \hline
\end{tabular}
\label{tab:activitynet}
\vspace{-1mm}
\end{table}

\begin{table}
\centering
\small
\caption{\small Performance comparisons on CUB-200-2011 in two scenarios: where the object bounding boxes are provided at training and test time or not. ft: fine-tuning.}
\begin{tabular}{|@{~}l@{~}|@{~}c@{~}|c|c|c|c|} \hline
Methods                  & dim     & ~~w/o ft~~    & w ft      & +box w/o ft    & +box w ft \\ \hline \hline
GA               & 4k    & 61.0\%    & 70.4\%    & 65.3\%        & 76.4\% \\
FV               & 128k   & 70.8\%    & 74.0\%    & 73.6\%        & 77.1\% \\
VLAD             & 128k  & 73.5\%    & 76.5\%    & 75.1\%        & 79.8\% \\
BP              & 256k  & 75.2\%    & 78.0\%    & 76.9\%        & 80.8\% \\ \hline
FV-VAE             & 128k  & 79.3\%    & 82.4\%    & 79.5\%        & 83.6\% \\ \hline
\multicolumn{2}{|p{0.20\columnwidth}|}{Previous work}    & \multicolumn{2}{p{0.32\columnwidth}|}{84.5\%\cite{zhang2016picking}~~84.1\%\cite{lin2015bilinear} 84.1\%\cite{jaderberg2015spatial}~~82.0\%\cite{krause2015fine} 75.7\%\cite{branson2014bird}~~~~73.9\%\cite{zhang2014part}} & \multicolumn{2}{p{0.32\columnwidth}|}{85.1\%\cite{lin2015bilinear}~~82.8\%\cite{krause2015fine} 76.4\%\cite{zhang2014part}~~73.0\%\cite{cimpoi2016deep}}  \\ \hline
\end{tabular}
\label{tab:cub}
\vspace{-4mm}
\end{table}

The results across different evaluation metrics consistently indicate that visual representation produced by our FV-VAE leads to a performance boost against baselines on ActivityNet validation set, as shown in Table \ref{tab:activitynet}. More specifically, FV-VAE on the outputs of pool5 in VGG\_19 and res5c in ResNet\_152 outperforms GA from VGG\_19 and ResNet\_152 by 10.0\% and 9.8\% in terms of mAP, respectively. Furthermore, the representation learnt by FV-VAE only on visual appearance of video frame also exhibits better performance than GA representation from C3D and IDT motion features which additionally explore temporal information in videos.

\begin{figure*}[!tb]
   \centering
   \subfigure[]{
     \label{fig:fig4:a}
     \includegraphics[width=0.3\textwidth]{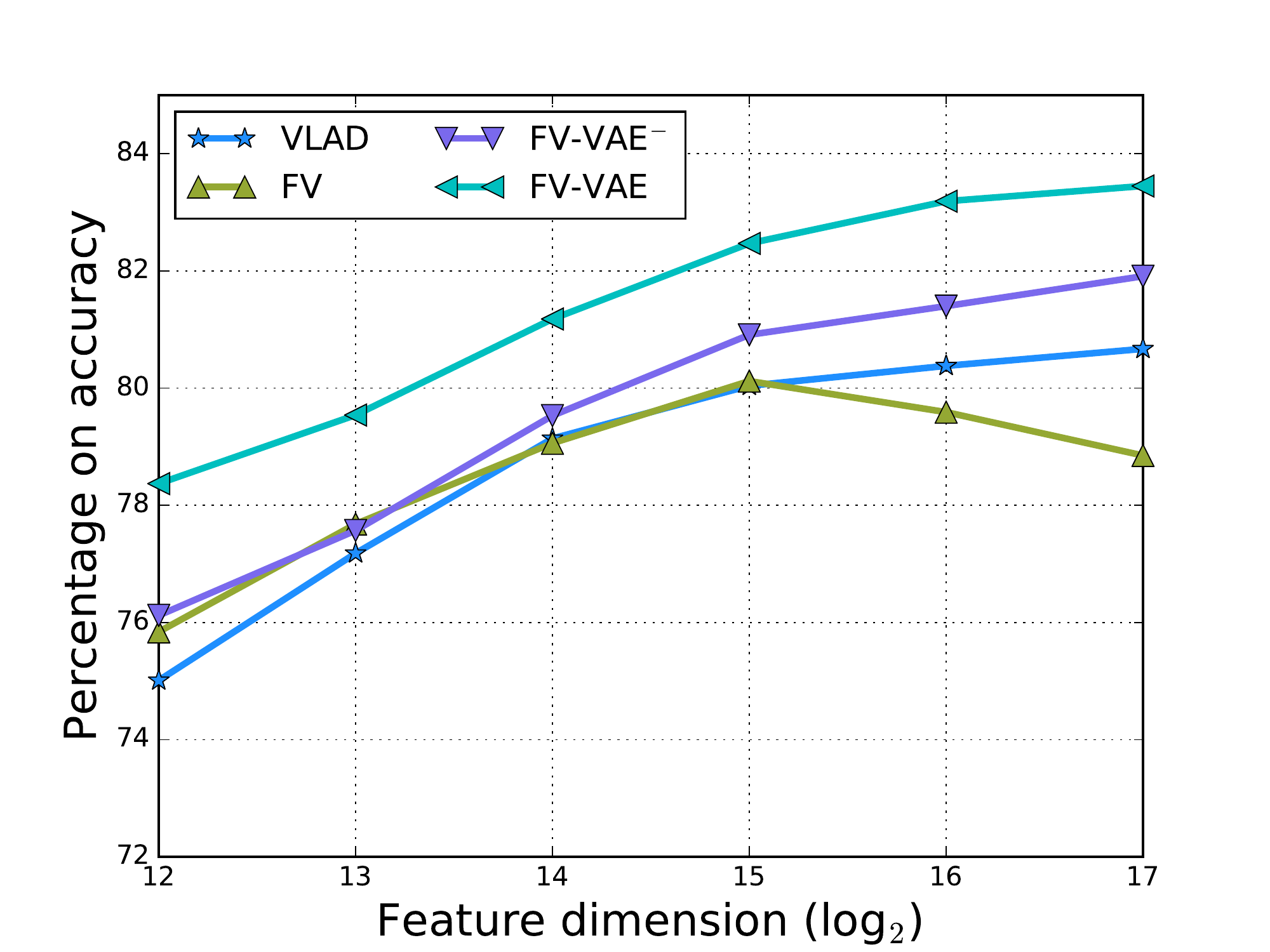}}
   \subfigure[]{
     \label{fig:fig4:b}
     \includegraphics[width=0.3\textwidth]{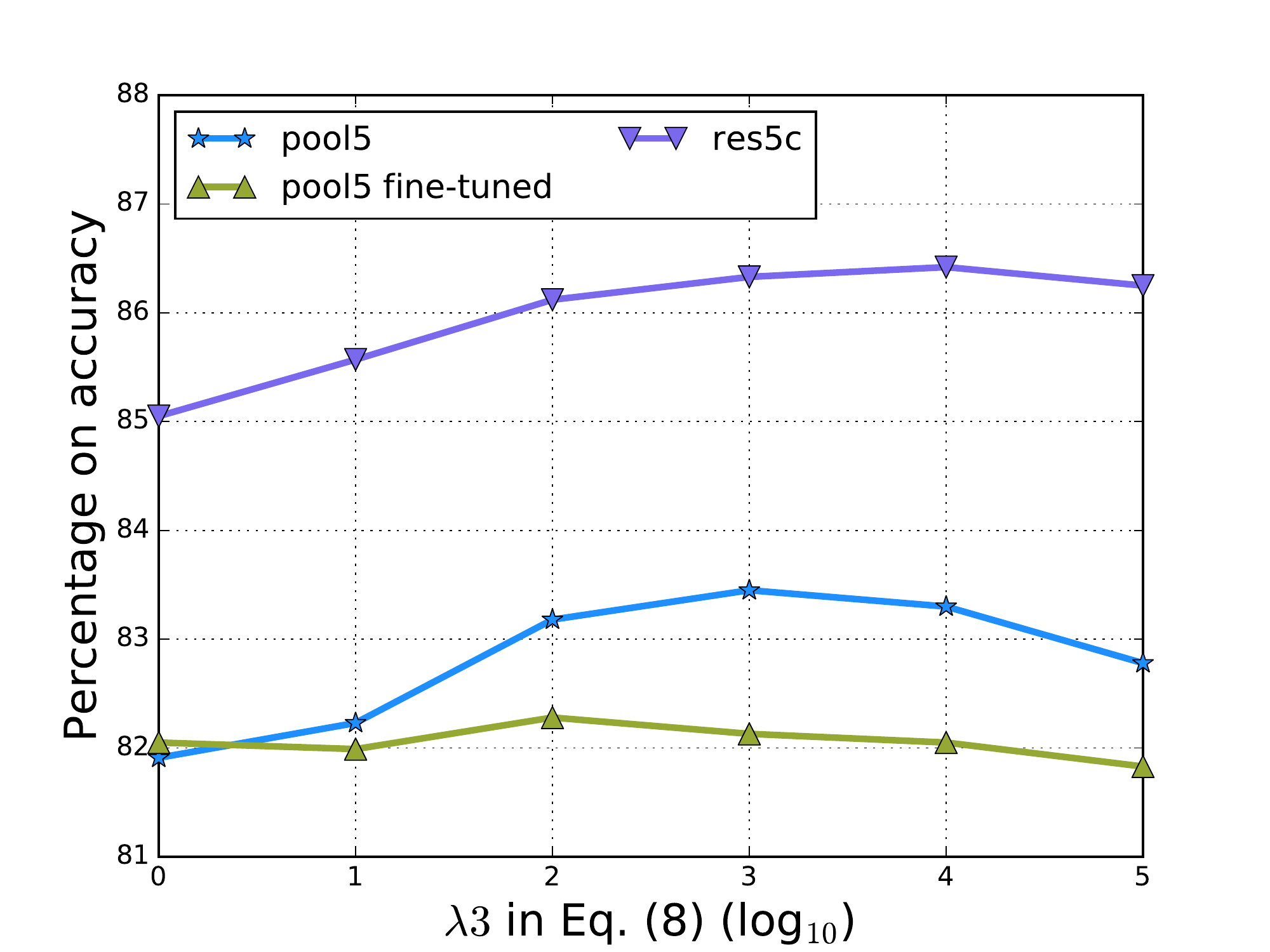}}
   \subfigure[]{
     \label{fig:fig4:c}
     \includegraphics[width=0.3\textwidth]{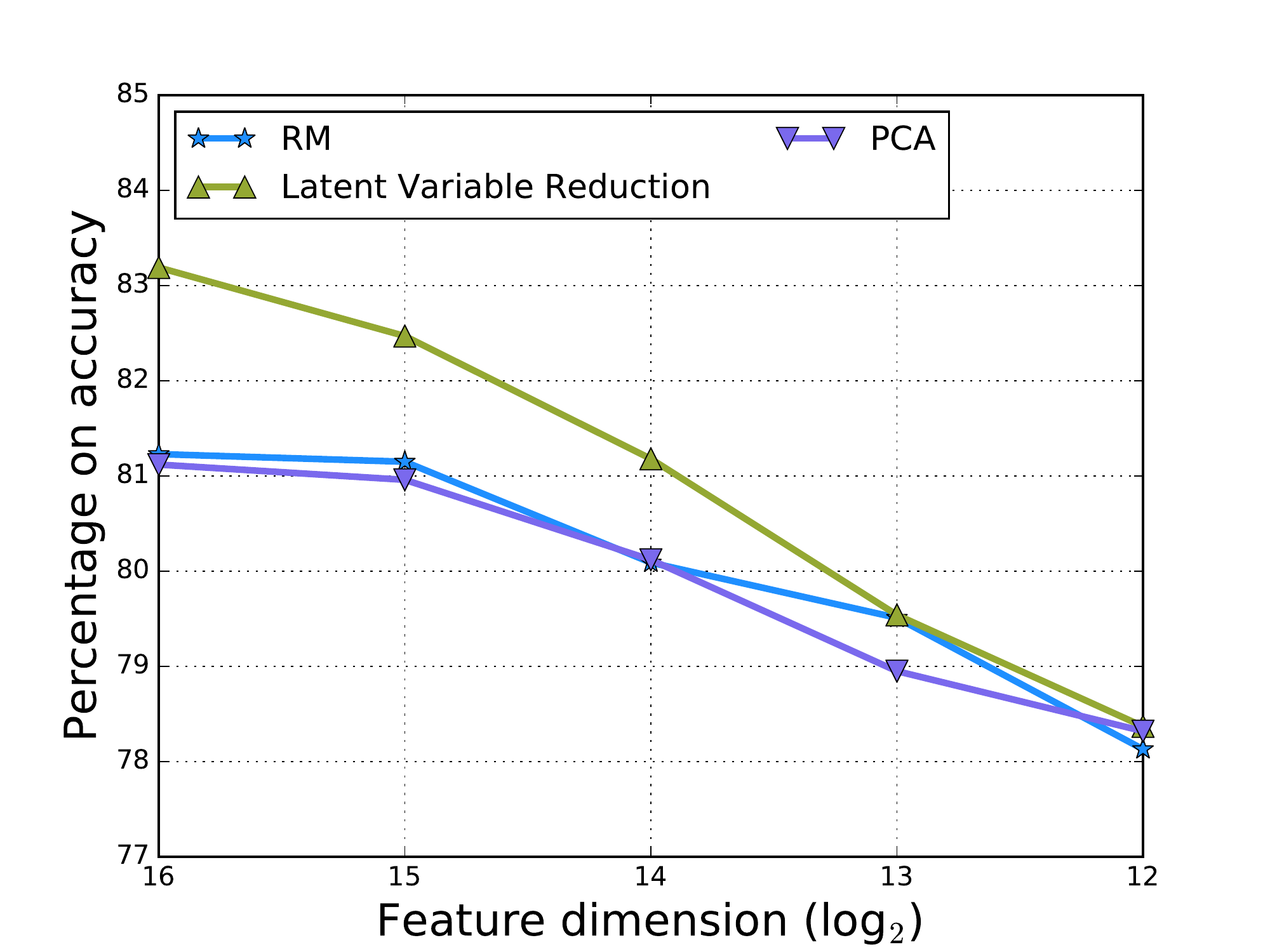}}
   \caption{\small Experimental analysis: (a) The accuracy of visual representation with different dimensions learnt by different quantization methods. (b) The accuracy curve of FV-VAE on activations from different networks with different ${\lambda}_{3}$ in Eq.(\ref{equ:LF}). (c) The accuracy of different feature compression methods on representation learnt by FV-VAE. Note that all the performances reported in this figure are on UCF101 split1 and similar performance trends are observed at the other two datasets.}
   \label{fig:fig4}
   \vspace{-2.5mm}
\end{figure*}

Fine-tuning VGG\_19 on CUB-200-2011 for FV-VAE generally performs better than original VGG\_19 on both protocols of where the object bounding boxes are given or not, as shown in Table \ref{tab:cub}. Overall, the representation learnt by FV-VAE leads to a performance boost against some baselines, e.g., \cite{krause2015fine} which extracts representation on local regions learnt by co-segmentation and \cite{branson2014bird} which combines the representations from three networks fed by warped bird head, warped body and entire image, respectively. It is not surprise that FV-VAE yields inferior performance to the other baselines, as the representation learnt by our FV-VAE is for general purpose while contributions of different regions in particular for fine-grained classification are taken into account in these methods. For instance, a saliency weight is learnt and assigned to each local region in \cite{zhang2016picking}, and a spatial transformer is trained to reduce the effect of translation and rotation as preprocess in \cite{jaderberg2015spatial}. More importantly, the importance estimation of each local region can be easily integrated into our framework as spatial attention.

\subsection{Experimental Analysis}
\textbf{The effect of representation dimension.} Figure \ref{fig:fig4:a} compares the accuracy of learnt representations with different dimensions by changing the number of latent variable in FV-VAE, the number of centroids in VLAD and mixture components in FV. Overall, visual representation learnt by FV-VAE consistently outperforms others at each dimension from $2^{12}$ to $2^{17}$. In general, higher dimensional representations provide better accuracy, except that the accuracy of representation learnt by FV will decrease when the dimension is higher than $2^{15}$, which may caused by overfitting. The result basically indicates the advantage of predicting Gaussian parameters by a neural network in our FV-VAE.

\textbf{The effect of tradeoff parameter ${\lambda}_{3}$.} A common problem with combination of multiple loss is the need to set the tradeoff parameters in between. Figure~\ref{fig:fig4:b} shows the accuracy of FV-VAE with respect to different ${\lambda}_{3}$ in Eq.(\ref{equ:LF}), which reflects the contribution of leveraging semantic information. As expected, the accuracy curves are all like the ``$\wedge$" shapes when ${\lambda}_{3}$ varies from $10^{0}$ to $10^{5}$.

\textbf{Feature compression.} Figure \ref{fig:fig4:c} compares the performance obtained by applying different representation compression methods: (1) Random Maclaurin (RM) \cite{kar2012random}, (2) PCA dimension reduction and (3) reducing the number of latent variable in VAE. Compared to RM and PCA which separately learn a transformation for feature compression, we can reduce the dimension of the learnt FV in VAE framework by decreasing the number of latent variable. As indicated by our results, reducing the number of latent variable always achieves the best accuracy, which again confirms the high flexibility of VAE.

\section{Conclusion}
We have presented Fisher Vector with Variational Auto-Encoder (FV-VAE) architecture which aims to quantize the convolutional activations in a deep generative model. Particularly, we theoretically formulate the computation of FV in VAE architecture. To verify our claim, a general visual representation learning framework is devised by integrating our FV-VAE architecture and an implementation of FV-VAE is also substantiated for image and video recognition. Experiments conducted on on three public datasets, i.e., UCF101, ActivityNet, and CUB-200-2011 in the context of video action recognition and fine-grained image classification validate our proposal and analysis. Performance improvements are clearly observed when comparing to other quantization techniques.

Our future works are as follows. First, a deeper auto-encoder architecture will be explored in our FV-VAE architecture. Second, attention mechanism will be explicitly incorporated into our FV-VAE for further enhancing visual recognition. Third, Generative Adversarial Networks (GAN) will be investigated to better learn a generative model and integrated into representation learning.

\section*{Appendix}
\begin{table*}[!tb]
\centering
\small
\caption{\small The detailed architecture of our FV-VAE. The operation, the spatial dimension and the number of output channels are given for each layer. Layers 1$\sim$3 are pre-processing steps for the purpose of stable optimization. Layers 4$\sim$5 are ``encoder'' and layer 7 is ``decoder'' in VAE.}
\begin{tabular}{|l|*{8}{c|}} \hline
Task&   input         & 1 & 2 & 3 & 4 & 5 & 6 & 7   \\ \hline \hline
video action & \multirow{2}{*}{\textbf{pool5}} & \multirow{2}{*}{\textbf{L2 norm}} & \multirow{2}{*}{\textbf{spp}} & \multirow{2}{*}{\textbf{dropout}} & \multirow{2}{*}{1$\times$1\textbf{conv}+\textbf{relu}} & \multirow{2}{*}{1$\times$1\textbf{conv}} & \multirow{2}{*}{\textbf{sampling}} & \multirow{2}{*}{1$\times$1\textbf{conv}+\textbf{relu}}\\
recognition & &&&&&&&\\ \hline
\multicolumn{1}{|r|}{\emph{--- spatial dim}} & 7$\times$7  & 7$\times$7 & 1$\times$50 & 1$\times$50 & 1$\times$50 & 1$\times$50 & 1$\times$50 & 1$\times$50 \\
\multicolumn{1}{|r|}{\emph{--- \#channel~~~~}} & 512  & 512 & 512 & 512 & 512 & 256 & 256 & 512 \\ \hline \hline
video action & \multirow{2}{*}{\textbf{res5c}} & \multirow{2}{*}{\textbf{L2 norm}} & \multirow{2}{*}{\textbf{spp}} & \multirow{2}{*}{\textbf{dropout}} & \multirow{2}{*}{1$\times$1\textbf{conv}+\textbf{relu}} & \multirow{2}{*}{1$\times$1\textbf{conv}} & \multirow{2}{*}{\textbf{sampling}} & \multirow{2}{*}{1$\times$1\textbf{conv}+\textbf{relu}}\\
recognition & &&&&&&&\\ \hline
\multicolumn{1}{|r|}{\emph{--- spatial dim}} & 7$\times$7  & 7$\times$7 & 1$\times$50 & 1$\times$50 & 1$\times$50 & 1$\times$50 & 1$\times$50 & 1$\times$50 \\
\multicolumn{1}{|r|}{\emph{--- \#channel~~~~}} & 2048  & 2048 & 2048 & 2048 & 2048 & 256 & 256 & 2048 \\ \hline \hline
fine-grained & \multirow{2}{*}{\textbf{conv$_{5\_4}$+relu}} & \multirow{2}{*}{\textbf{L2 norm}} & \multirow{2}{*}{--} & \multirow{2}{*}{\textbf{dropout}} &  \multirow{2}{*}{1$\times$1\textbf{conv}+\textbf{relu}} & \multirow{2}{*}{1$\times$1\textbf{conv}} & \multirow{2}{*}{\textbf{sampling}} & \multirow{2}{*}{1$\times$1\textbf{conv}+\textbf{relu}}\\
image classification & &&&&&&&\\ \hline
\multicolumn{1}{|r|}{\emph{--- spatial dim}} & 28$\times$28  & 28$\times$28 & -- & 28$\times$28 & 28$\times$28 & 28$\times$28 & 28$\times$28 & 28$\times$28 \\
\multicolumn{1}{|r|}{\emph{--- \#channel~~~~}} & 512  & 512 & -- & 512 & 512 & 256 & 256 & 512 \\ \hline
\end{tabular}
\label{tab:arch}
\vspace{-0mm}
\end{table*}

\begin{figure*}[!tb]
   \centering {\includegraphics[width=0.95\textwidth]{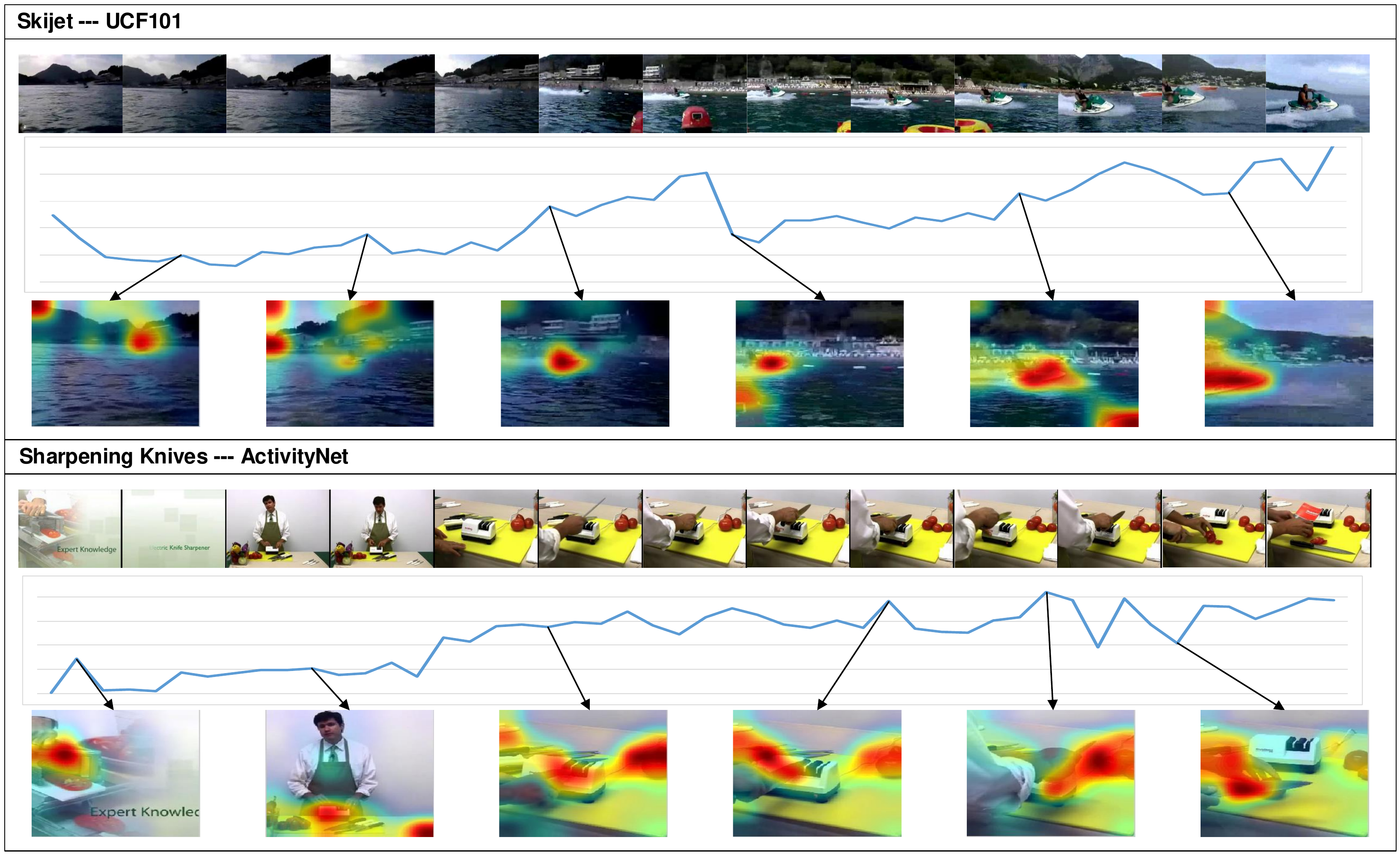}}
   \caption{\small Feature visualization from the viewpoint of spatial and temporal attention for one video example in UCF101 and ActivityNet dataset respectively (Top row: sampled frames from the video, middle row: the curve of frame-level values, bottom row: region-level values in each frame). The sum of the absolute value of each element in the FV of one regional descriptor represents spatial attention of this region and then the sum of the value of each region in one frame is considered as temporal attention of that frame. We can see that it is able to concentrate attention to regions of interest, e.g., jet boat in the first video and sharpener in the second one, which highly infer the action ``Skijet" and ``Sharpening Knives" happening in the video. Moreover, it is also capable of predicting the contributions of different temporal frames. For example, frames with motion of ``a man is riding a jet boat" in the first video or ``a man is sharpening a knife" in the second video contribute more to the recognition.}
   \label{fig:fig5}
   \vspace{-3mm}
\end{figure*}

\subsection*{A. Network Architecture}

Table \ref{tab:arch} details the hyper-parameters of our proposed FV-VAE. Specifically, three different architectures are listed on different tasks and convolutional activations from different layers, i.e., video action recognition on activations of pool5 layer in VGG\_19 \cite{Simonyan:ICLR15} and res5c layer in ResNet\_152 \cite{he2015deep}, and fine-grained image classification on activations of conv$_{5\_4}$+relu layer in VGG\_19.

\begin{figure*}[!tb]
   \centering {\includegraphics[width=0.92\textwidth]{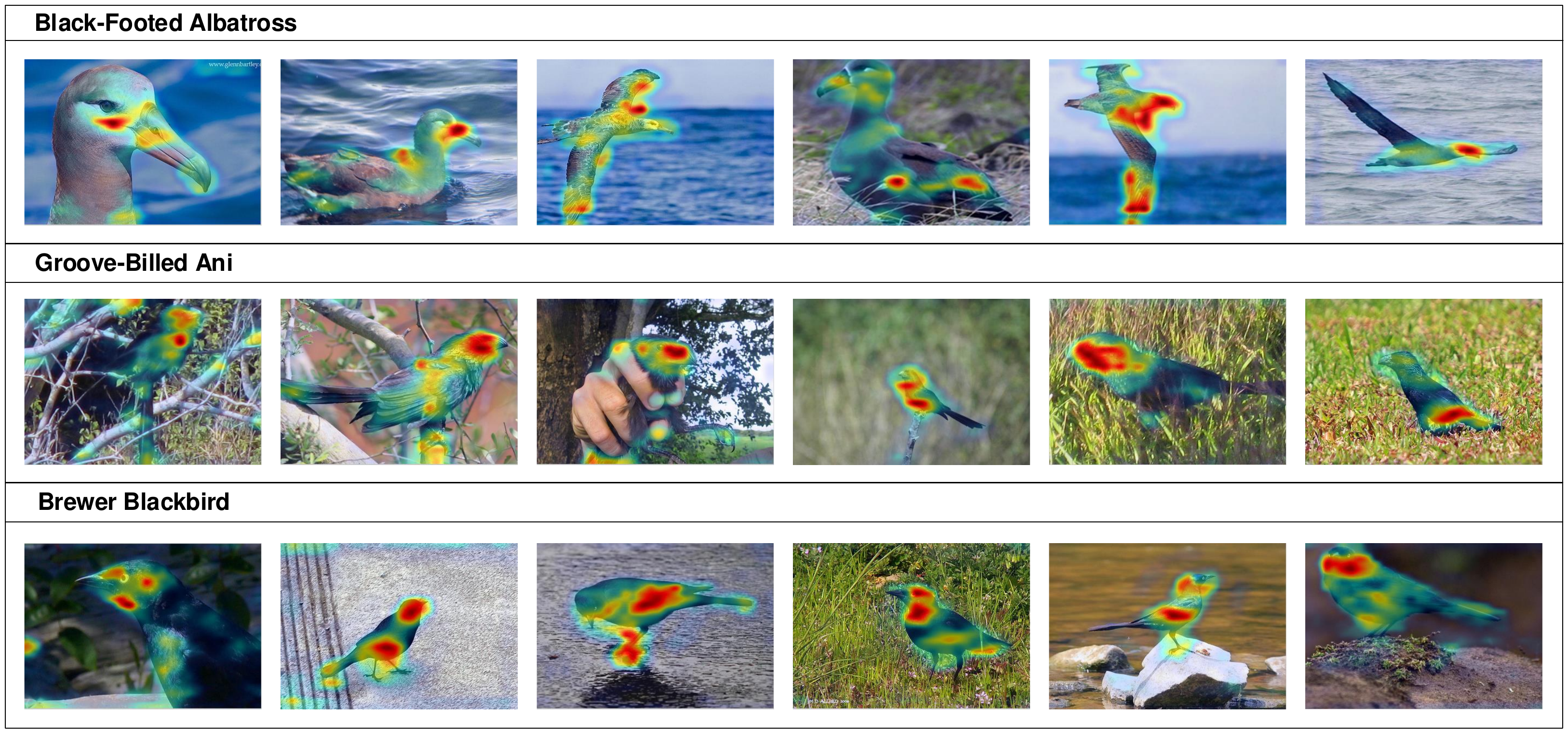}}
   \caption{\small Feature visualization from the viewpoint of spatial attention for image examples in CUB-200-2011 dataset. The sum of the absolute value of each element in the FV of one regional descriptor represents spatial attention of this region. We can see that most of the high-value regions are beak, neck and wing of birds, where the parts highly indicate the category that a bird belongs to.}
   \label{fig:fig6}
   \vspace{-6mm}
\end{figure*}

\subsection*{B. Feature Visualization}

Recall that FV of each regional descriptor is the gradient vector of reconstruction loss, as discussed in Section \ref{sec:FVEX}. The absolute value of each element in the vector in fact reflects its degree of being reconstructed and thus the sum of the absolute value of each element (region-level value) represents the importance of this region to recognition. The importance map can be regarded as spatial attention. Furthermore, the sum of the value of each region in one video frame (frame-level value) then manifests its score in the whole video sequence, which can be considered as temporal attention.

Figure \ref{fig:fig5} illustrates both region-level and frame-level values on one video example from ``Skijet'' and ``Sharpening Knives" category in UCF101~\cite{UCF101} test set and ActivityNet \cite{caba2015activitynet} validation set, respectively. The video is represented by sampled frames in the top row. The curve of frame-level score is given in the middle row. The scores of temporal frames with clear motion of ``a man is riding a jet boat" in the first video or ``a man is sharpening a knife" in the second video are high, indicating that these frames contribute more to the recognition. In the bottom row, the regions of high values are around the object jet boat or sharpener, which is the key evidence of ``Skijet" or ``Sharpening Knives" category, respectively.

Similar in spirit, Figure \ref{fig:fig6} visualizes region-level values of a few exemplary images in CUB-200-2011~\cite{WahCUB_200_2011} test set. The image examples are from three bird species, i.e., ``Black-Footed Albatross,'' ``Groove-Billed Ani'' and ``Brewer Blackbird.'' We can easily observe that most of the high-value regions are beak, neck and wing of birds, where the parts are found to be more helpful to differentiate one variety of bird from the other. In contrast, the regions around background (e.g., sea, grass, branches, rocks) receive low values.

{\small
\bibliographystyle{ieee}
\bibliography{egbib}
}

\end{document}

%% file: commands.tex
\usepackage{amsthm}
\usepackage{amsmath}
\usepackage{amsfonts}
\usepackage{amssymb}

\newcommand{\bb}[1]{\mathbf{#1}}

\newcommand{\bx}{\bb{x}}
\newcommand{\bxt}{\bx_{t}}
\newcommand{\by}{\bb{y}}
\newcommand{\byt}{\by_{t}}
\newcommand{\bz}{\bb{z}}
\newcommand{\bzt}{\bz_{t}}

\newcommand{\bTheta}{\boldsymbol{\theta}}
\newcommand{\bPhi}{\boldsymbol{\phi}}
\newcommand{\bMu}{\boldsymbol{\mu}}
\newcommand{\bSigma}{\boldsymbol{\sigma}}
\newcommand{\bEpsilon}{\boldsymbol{\epsilon}}

\newcommand{\uTheta}{u_{\bTheta}}
\newcommand{\half}{\frac{1}{2}}

\newcommand{\pTheta}{p_{\bTheta}}
\newcommand{\qPhi}{q_{\bPhi}}